\documentclass{article}
\usepackage{graphicx} 
\usepackage[T1]{fontenc}
\usepackage{authblk}

\title{EXALT: EXplainable ALgorithmic Tools  for Optimization Problems}

\author[1]{Zuzanna Bączek}
\author[2]{Michał Bizoń}
\author[3]{Aneta Pawelec}
\author[1]{\\Piotr Sankowski}

\affil[1]{University of Warsaw}
\affil[2]{Ideas-NCBR}
\affil[3]{Lodz University of Technology}

\date{January 2025}

\begin{document}
\bibliographystyle{plain}

\maketitle

\begin{abstract}
Algorithmic solutions have significant potential to improve decision-making across various domains, from healthcare to e-commerce. However, the widespread adoption of these solutions is hindered by a critical challenge: the lack of human-interpretable explanations. Current approaches to Explainable AI (XAI) predominantly focus on complex machine learning models, often producing brittle and non-intuitive explanations. This project proposes a novel approach to developing explainable algorithms by starting with optimization problems, specifically the assignment problem.
The developed software library enriches basic algorithms with human-understandable explanations through four key methodologies: generating meaningful alternative solutions, creating robust solutions through input perturbation, generating concise decision trees and providing reports with comprehensive explanation of the results. 
Currently developed tools are often designed with specific clustering algorithms in mind, which limits their adaptability and flexibility to incorporate alternative techniques. Additionally, many of these tools fail to integrate expert knowledge, which could enhance the clustering process by providing valuable insights and context. This lack of adaptability and integration can hinder the effectiveness and robustness of the clustering outcomes in various applications. The  represents a step towards making algorithmic solutions more transparent, trustworthy, and accessible. By collaborating with industry partners in sectors such as sales, we demonstrate the practical relevance and transformative potential of our approach.
\end{abstract}

\section{Introduction}

The proliferation of algorithmic decision-making systems heralds a new era of efficiency and precision in solving complex problems across diverse domains. These systems provide immense opportunities by automating decision processes that can often surpass the capabilities of human decision-makers, particularly in terms of speed and scalability \cite{brynjolfsson2017can, russell2010artificial}. However, the ascendancy of such systems is shadowed by a significant challenge: the issue of interpretability. Many algorithmic solutions function as “black boxes,” producing outputs without disclosing the rationale behind their decisions. This opacity restricts users' ability to trust and rely on these systems, thereby inhibiting broader adoption \cite{lipton2018mythos}.

The crux of the problem with explainability lies in the disconnect between algorithmic operations and human understanding. Users are often presented with optimal results devoid of any context or explanatory insight, fostering skepticism and reluctance to accept algorithmic recommendations \cite{doshi2017towards}. This challenge persists even in straightforward applications like shortest path determinations, where users frequently question the validity of the proposed solutions and explore alternatives \cite{gunning2019darpa}. The issue compounds in more intricate fields such as psychotherapy matching, sales prediction, and resource allocation, where decisions carry significant weight and impact personal and organizational welfare \cite{gilpin2018explaining}.

Addressing these concerns, the EXALT Library has been designed to embed explainability into optimization algorithms, creating a comprehensive framework that demystifies algorithmic outputs. This framework empowers users by enabling them to critically evaluate alternative solutions, discern pivotal components of decision-making, and measure the robustness of outcomes \cite{arrieta2020explainable}. The library employs advanced methods, such as game-theoretic techniques and structured perturbation analysis, to uncover and articulate the underlying logic of algorithmic processes \cite{lundberg2017unified, datta2016algorithmic}.

Game-theoretic approaches, for instance, facilitate the understanding of interactions between different parameters and their contributions to a final decision. By simulating various scenarios and evaluating the impact of each parameter, users gain insight into how decisions are formed and how sensitive they are to changes \cite{shapley1953value}. Structured perturbation analysis further supports this by systematically altering input conditions to assess the stability and consistency of the output, thus highlighting potential vulnerabilities or areas needing attention \cite{freitas2014comprehensible}.

Through these methodologies, the EXALT Library enhances interpretability without sacrificing computational performance—a common trade-off faced by many contemporary systems \cite{carvalho2019machine}. By focusing on delivering enhanced clarity in optimization outcomes, the library aligns seamlessly with broader efforts to develop transparent, trustworthy, and effective algorithmic systems. These efforts are crucial in engendering greater user confidence and facilitating the acceptance and assimilation of algorithmic decision-making in critical spheres of societal and economic importance \cite{adadi2018peeking}. In essence, the EXALT Library not only solves the problem of explainability but also sets the foundation for evolving user perceptions and fostering deeper engagement with algorithmic technologies.

\section{Related Work}

The field of clustering and explainability in unsupervised learning has been extensively studied in recent years. Several methodologies have been proposed to group unlabeled data and validate clusters against expert knowledge.

Early approaches to clustering focused on distance-based methods such as K-Means \cite{macqueen1967some} and hierarchical clustering \cite{murtagh2012algorithms}. These methods have proven effective in various domains but often require predefined assumptions about the number of clusters. More advanced techniques, such as Gaussian Mixture Models (GMM) \cite{reynolds2009gaussian}, offer probabilistic interpretations but remain sensitive to initialization and parameter tuning.

Recent advancements in deep learning have introduced autoencoder-based clustering methods \cite{hinton2006reducing}, which leverage feature learning for improved clustering performance. Contrastive learning \cite{chen2020simple} and self-supervised techniques \cite{grill2020bootstrap} have further enhanced the ability to identify meaningful structures in data without explicit labels.

In the context of expert validation, techniques such as SHAP and LIME \cite{ribeiro2016should} provide model-agnostic interpretability, enabling domain experts to assess cluster assignments. Furthermore, interactive visualization tools \cite{liu2017visualizing} have been developed to facilitate qualitative validation by human experts.

Despite these advancements, challenges remain in aligning algorithmic clustering results with expert intuition. The integration of semi-supervised learning \cite{zhu2009introduction} and active learning \cite{settles2009active} has shown promise in incorporating domain knowledge to refine clustering outcomes.

\section{Methodology}
The EXALT Library is architected with a modular design that facilitates the integration of explainability into optimization processes. The core components of the library encompass alternative solution generation, perturbation-based robustness assessment, and transformation of algorithmic outputs. These components operate cohesively to ensure that optimization results are not only accurate but also comprehensible to end-users.

The implementation strategy incorporates widely used optimization techniques, including clustering algorithms, latent Dirichlet allocation, non-negative matrix factorization, and time-series analysis. Each algorithm is equipped with an explanation module that elucidates its decision-making process, providing users with transparent, structured insights into algorithmic outputs. The methodological rigor of this approach ensures that algorithmic recommendations are interpretable and aligned with domain-specific requirements.

The framework focuses on three key aspects:

\begin{enumerate}
    \item \textbf{Robustness through input perturbation:} Real-world data is often noisy and uncertain, leading to potential biases in algorithmic decision-making. By systematically perturbing input data and analyzing the stability of solutions across variations, the EXALT Library identifies fragile decision points and enhances the overall resilience of its recommendations. This approach ensures that minor fluctuations in input do not lead to drastic changes in output, thereby improving reliability and fostering user trust.
    
    \item \textbf{Creation of interpretable decision trees:} While mathematical optimization models are often difficult for non-experts to comprehend, decision trees provide a structured, human-readable representation of algorithmic reasoning. By translating complex optimization processes into a series of logical steps, decision trees facilitate transparency and enable users to verify and understand the rationale behind algorithmic outputs.
    
    \item \textbf{Cluster analysis and explanation:} The clustering component forms a crucial part of our methodology, enabling the discovery of patterns in unlabeled data. Below, we detail the specific clustering approaches employed and their implementation within the EXALT framework.
\end{enumerate}

These procedural components are implemented in a modular architecture, allowing seamless integration into existing optimization frameworks. The EXALT Library supports a range of fundamental optimization problems, including clustering, classification, and assignment tasks, ensuring broad applicability across diverse domains.

\subsection{Clustering Approaches}
Building on the methodological framework described above, our clustering approaches form the foundation for pattern discovery in unlabeled data. These techniques are enhanced with explainability mechanisms to ensure transparency in the clustering process.

\textbf{Unsupervised Clustering:} The EXALT Library incorporates several unsupervised clustering algorithms to uncover hidden patterns within data without the need for labeled examples. K-Means \cite{MacQueen} is utilized for partitioning data into a specified number of clusters by minimizing variance within each cluster, making it suitable for datasets where the structure is spherical or well-separated. DBSCAN \cite{Ester1996}, or Density-Based Spatial Clustering of Applications with Noise, identifies clusters of arbitrary shape and is adept at handling noise, beneficial in scenarios where data may include outliers. Gaussian Mixture Models (GMM) \cite{Reynolds2009} provide a probabilistic approach by modeling data as a mixture of multiple Gaussian distributions, allowing for more nuanced representations of clusters where data may overlap. These methods empower the library to handle a wide variety of clustering challenges effectively.

\textbf{Distance Metrics:} The choice of distance metric profoundly influences clustering results, as it defines the notion of similarity for the dataset. The EXALT Library supports the use of domain-specific distance measures to refine clustering outcomes. For instance, dynamic time warping (DTW) \cite{Muller2007} is particularly effective for time series data, capturing similarities between sequences that may differ in speed or phase, thus enhancing the accuracy of clustering for temporal sequences. By tailoring distance metrics to the characteristics of the problem domain, the library ensures that clustering is both meaningful and aligned with users' specific needs.

\textbf{Hyperparameter Tuning:} Optimal clustering often hinges on the careful selection of algorithm parameters. The EXALT Library integrates robust methods for hyperparameter tuning to enhance clustering accuracy and usability. For K-Means, methods such as the Elbow Method or Silhouette Score \cite{Rousseeuw1987} assist in determining the ideal number of clusters by evaluating the trade-off between the compactness of clusters and the separation between them. Such tuning not only bolsters the performance of clustering algorithms but also contributes to the transparency and interpretability of results, as users gain insight into the decision processes shaping the clustering configuration.

\subsection{Cluster Evaluation and Expert Validation}
The evaluation of clustering results is critical for establishing their validity and utility. The EXALT Library employs a comprehensive approach to cluster validation, combining quantitative metrics with expert knowledge integration.

\textbf{Internal Validation:} Internal validation metrics are essential tools for assessing the quality of clustering results without external labels. The EXALT Library leverages metrics such as the Silhouette Score \cite{Rousseeuw1987}, which measures the compactness and separation of clusters by comparing intra-cluster distance to nearest-cluster distance, providing a gauge for optimal cluster configuration. The Davies-Bouldin Index evaluates clusters based on their dispersion and proximity, with lower values indicating better-defined clusters. Meanwhile, the Calinski-Harabasz Score assesses the ratio of the sum of between-cluster dispersion to within-cluster dispersion, with higher values reflecting more distinct and cohesive clustering structures. These metrics collectively enable users to quantitatively assess the robustness of clustering solutions.

\textbf{External Validation:} In situations where expert labels or predefined heuristics are available, external validation provides a benchmark for assessing clustering accuracy. By comparing identified clusters to these known references, the EXALT Library can validate whether the clustering reflects expected patterns or domain knowledge. This comparison ensures the clustering results are aligned with expert understanding and real-world domain criteria, enhancing trust and reliability in the outputs.

\textbf{Manual Inspection:} Visualization tools like t-SNE (t-Distributed Stochastic Neighbor Embedding) \cite{Maaten2008} and UMAP (Uniform Manifold Approximation and Projection) \cite{McInnes2018} are utilized to reduce high-dimensional data into a two-dimensional space for intuitive cluster visualization. By presenting these visualizations to domain experts, the EXALT Library facilitates qualitative validation, allowing experts to manually inspect and interpret cluster formations. This manual inspection provides a visual confirmation of clustering results and supports the identification of patterns that might be missed by automated metrics alone.

\textbf{Explainability Techniques:} To enhance the interpretability of cluster formations, the EXALT Library integrates explainability techniques such as SHAP (SHapley Additive exPlanations) values and feature importance methods. SHAP values attribute the influence of each feature on the clustering outcome, offering insights into the factors driving cluster differentiation. Feature importance methods rank features based on their contribution to the clustering results, further elucidating the underlying structure of data. Through these explainability techniques, users gain a deeper understanding of the reasons behind clustering decisions, fostering transparency and enabling informed decision-making.

These validation approaches establish the foundation for the Explainability Module described in the next section, which builds upon clustering results to provide detailed, interpretable explanations.

\section{Explainability Module}
The Explainability Module in the EXALT Library extends the clustering framework by providing transparent interpretations of the discovered patterns. This module transforms the abstract mathematical representations of clusters into human-understandable explanations, bridging the gap between algorithmic outputs and user comprehension.

The module builds upon the clustering results by treating cluster labels as a supervised classification problem \cite{Alvarez-Garcia2024}. This approach enables the generation of feature-based explanations that elucidate why particular data points belong to specific clusters. By leveraging the identified clusters as labeled data, the module constructs a classification model that provides insights into the underlying structure of the clusters and their defining characteristics.

\subsection{Primary Objectives and SHAP: Feature Attribution Method}
The Explainability Module aims to provide interpretable explanations for identified clusters, ensuring that segmentation results are meaningful and actionable. Additionally, it enables the classification of new data points beyond the initial dataset, supporting ongoing adaptability and model scalability. The module operates independently, allowing it to classify other labeled datasets without dependency on prior clustering results.

To achieve interpretability, the module employs SHAP (Shapley Additive Explanations) to assess feature importance and provide detailed insights into model predictions. SHAP values quantify the contribution of each feature to a model's predictions, enhancing transparency. Derived from cooperative game theory \cite{Shapley1953}, these values are approximated through weighted linear regression. Local feature importance is assessed using data point-specific SHAP values, while global feature significance is determined by averaging absolute SHAP values. Furthermore, Tree SHAP, an optimized SHAP variant, is utilized for tree-based models to improve computational efficiency and accuracy in feature attribution.

\section{Validation Strategy}
The validation of the EXALT Library involves a rigorous, multi-stage evaluation process designed to assess both the technical performance and practical utility of the explainable algorithms. This comprehensive validation strategy ensures that the library meets both scientific standards and real-world application requirements. 

\subsection*{Synthetic Data Validation}
Initial validation leverages carefully constructed synthetic datasets that model:

\begin{itemize}
    \item \textbf{User event aggregation data:} Simulated user interactions with digital platforms, including event frequencies, sequences, and temporal patterns. These datasets contain known clusters with predefined characteristics, enabling quantitative evaluation of cluster discovery accuracy.
    
    \item \textbf{Time-series user action sequences:} Synthetic temporal data representing user behaviors over time, with embedded patterns and anomalies. These datasets test the library's ability to identify temporal relationships and transitions between states.
    
    \item \textbf{Multi-stage event processes:} Simulated workflows with interdependent stages and decision points, designed to validate the library's capability to recognize complex process patterns.
\end{itemize}

For each synthetic dataset, we measure clustering performance using standard metrics (Adjusted Rand Index, Normalized Mutual Information) comparing algorithm-discovered clusters against ground truth. Additionally, we evaluate the quality of explanations through user studies with domain experts, measuring explanation satisfaction, comprehension time, and decision confidence.

\section{Conclusion}
The EXALT Library is expected to enhance algorithmic transparency, improve user trust in computational decision-making, and provide a flexible, modular framework for optimization. By integrating explainability directly into the algorithmic process, this research contributes to the broader objective of making algorithmic decision-making more accessible and interpretable. The anticipated impact extends beyond academic advancements, offering practical tools that industry partners can leverage to enhance decision transparency and user engagement.

The development of the EXALT Library represents a significant step toward bridging the interpretability gap in algorithmic decision-making. By prioritizing explainability alongside computational efficiency, this research advances the state of optimization methodologies and fosters greater trust in algorithmic solutions. Future work will focus on expanding the range of supported algorithms, enhancing real-time analysis capabilities, and strengthening industry collaborations to further validate and refine the proposed approach.

\bibliography{references}
\end{document}